\documentclass{article}
\usepackage[final]{colm2026_conference}

\usepackage[T1]{fontenc}
\usepackage[utf8]{inputenc}

\usepackage{enumitem}
\usepackage{microtype}
\usepackage{hyperref}
\usepackage{url}
\usepackage{booktabs}
\usepackage{amsmath,amssymb}
\usepackage{graphicx}
\usepackage{xcolor}
\usepackage{tabularx}
\usepackage{threeparttable}
\usepackage{siunitx}
\usepackage{float}
\usepackage{wrapfig}
\usepackage{algorithm}
\usepackage{algpseudocode}

\usepackage{pifont}
\usepackage{multirow}
\usepackage{listings}
\lstdefinelanguage{PromptText}{%
  basicstyle=\ttfamily\scriptsize,%
  breaklines=true,%
  breakatwhitespace=false,%
  upquote=true,%
}
\newcommand{\cmark}{\textcolor{green!55!black}{\ding{51}}}%
\newcommand{\xmark}{\textcolor{red!70!black}{\ding{55}}}%

\definecolor{darkblue}{rgb}{0, 0, 0.5}
\hypersetup{colorlinks=true, citecolor=darkblue, linkcolor=darkblue, urlcolor=darkblue}
\title{FlowEvo: Self-Evolving Agents through the Co-Evolution of Workflows and Executable Skills}

\author{\bfseries Zeyu Ren$^{1}$\thanks{Equal contribution.} \quad Ling Yue$^{2}$\footnotemark[1] \quad Ran Li$^{3}$ \quad Yishu Wang$^{1}$ \quad Shengxiang Xu$^{1}$ \\
  \bfseries Hanmo Liu$^{3}$ \quad Shaowu Pan$^{2}$\thanks{Corresponding authors: \texttt{shimin.di@seu.edu.cn}, \texttt{pans2@rpi.edu}.} \quad Shimin Di$^{1}$\footnotemark[2] \\[4pt]
  \normalfont
  $^{1}$Southeast University \\
  $^{2}$Rensselaer Polytechnic Institute \\
  $^{3}$The Hong Kong University of Science and Technology
}

\begin{document}

\maketitle
\lhead{}
\renewcommand{\headrulewidth}{0pt}
\suppressfloats[t]

\begin{abstract}
Large language model agents can adapt to complex tasks by constructing workflows at inference time, but procedures discovered in one episode are usually discarded after execution. Existing skill libraries provide reusable executable routines, but are typically assembled offline and do not grow from the agent's own workflows. We introduce FlowEvo, a training-free framework in which workflows and skills co-evolve at inference time. FlowEvo compiles successful workflows into callable skills, stores them in a persistent bank, and uses retrieved skills either through direct execution or as context for constructing new workflows. It also tracks each skill's downstream utility and suppresses skills that cause negative transfer. Using a shared GPT-4o-mini backbone, FlowEvo achieves the highest accuracy among 8 baselines on the full standard splits of ALFWorld, HumanEval, MBPP, GSM8K, and MATH-500. On ALFWorld, it reaches 85.6\%, 26.4 points above the strongest baseline, while using roughly one third as many tokens. Across 10 base models spanning 7B to 671B parameters, FlowEvo outperforms ExpeL in 49 of 50 model--dataset comparisons. Code is available at \url{https://github.com/DEFENSE-SEU/FlowEvo}.
\end{abstract}

\section{Introduction}

\begin{figure}[t]
\centering
\includegraphics[width=\linewidth]{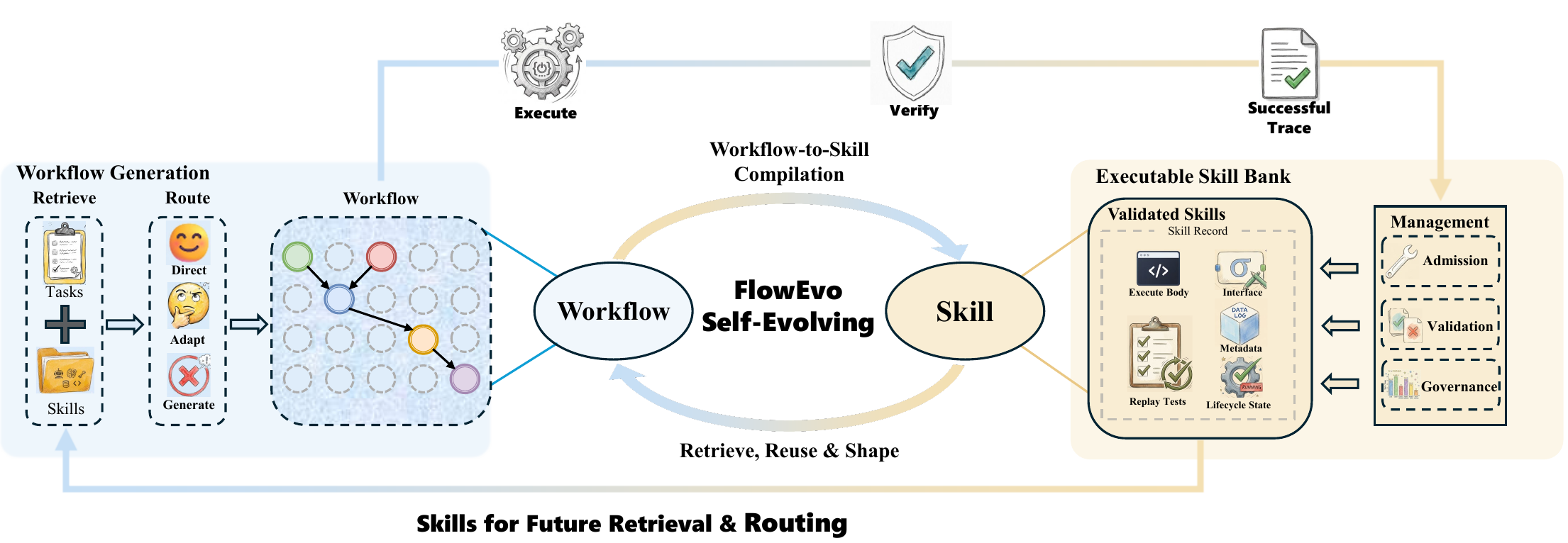}
\caption{Overview of FlowEvo. Successful workflows are compiled into executable skills and stored in a persistent bank. Retrieved skills can either be executed directly or supplied as structured context for subsequent workflow generation.}
\label{fig:overview}
\end{figure}

Large language model (LLM) agents increasingly solve tasks by constructing workflows at inference time rather than following a fixed prompt or a fixed sequence of tool calls \citep{hu2024adas,zhang2025aflow,wang2026learning,gao2025flowreasoner,wang2025scoreflow}. Solving a task may involve decomposition, tool use, code generation, intermediate checking, and repair \citep{wang2026peromas,xu2025robustflow,zheng2025mermaidflow}, and such agents are deployed across AI for science~\citep{xu2026ai}, from computational fluid dynamics~\citep{pan2026automating} and materials science~\citep{zhu2026survey} to Earth observation~\citep{yao2026remoteagent} and automated scientific peer review~\citep{wang2026ai}. This flexibility brings clear gains, but useful procedures discovered during one episode are rarely retained as \emph{skills}: executable routines, with an interface and usage guidance, that later tasks can invoke \citep{zhao2024expel}. Agents therefore repeatedly rediscover similar routines across related tasks, increasing token cost and variance \citep{sumers2023cognitive}. The reusable operational structure inside a successful workflow~\citep{zhu2023ghost}, such as a decomposition pattern or a stable code scaffold, is discarded with the transient trajectory of the episode that produced it.

Existing work addresses the two sides separately. One line stores prior plans, traces, or workflows as text and uses them to guide later generation \citep{wang2024awm,sarukkai2026self,zhang2026liveevo,zhang2025agentic,park2023generative}; a text record can describe what worked, but it cannot be invoked or validated, so reuse still depends on the model re-deriving the procedure. Another line treats tools, procedures, or skills as executable objects \citep{zheng2025skillweaver,chen2026cua,qian2023creator,wang2023voyager}; such libraries are callable and checkable, but they are typically constructed offline for a fixed task distribution, apart from the agent's own problem solving. Successful workflows lie between these two views. They contain more executable structure than text-based memory, but are less fixed than hand-designed tools. This raises a basic question: when an agent solves tasks through workflows, how should successful workflows be turned into reusable skills, and how should those skills affect later workflow construction when model parameters remain fixed?

We study this question with FlowEvo (Figure~\ref{fig:overview}), a training-free framework for self-evolving agents through the co-evolution of workflows and executable skills. When the agent solves a task successfully, the useful part of the workflow is compiled into a skill record and stored in a persistent bank; retrieved records are executed directly on compatible tasks and otherwise supplied as structured context for new workflows (Section~\ref{sec:method}).

This design creates a workflow--skill--workflow loop: execution produces successful traces, traces are compiled into executable skills, and the accumulated bank changes how later tasks are solved. The bank contributes in 2 ways: \emph{direct skill reuse}, where a validated callable artifact is executed on a new task, and \emph{skill-conditioned workflow generation}, where retrieved skills shape planning and generation without being replayed directly. Distinguishing the two matters because an improvement from skill accumulation should not be reduced to exact replay of previously stored solutions alone.

Experiments cover the full standard splits of 5 benchmarks, spanning an interactive environment (ALFWorld; \citealp{shridhar2020alfworld}) and code and math generation (HumanEval, MBPP, GSM8K, MATH-500). FlowEvo attains the best accuracy on every benchmark against 8 classic and recent baselines, the advantage holds across 10 base models, and controlled ablations attribute it to each mechanism (Section~\ref{sec:exp}).

Our contributions are as follows:
\begin{itemize}[leftmargin=*]
    \item \textbf{Workflow-to-skill compilation:} an online compilation step that turns successful workflows into reusable skill records combining a callable artifact with structured guidance.
    \item \textbf{Tiered skill-to-workflow reuse:} a routing mechanism over dynamic generation, direct skill execution, and skill-conditioned workflow generation, with direct execution falling back to conditioning when preconditions or validation fail.
    \item \textbf{Contrastive-utility lifecycle:} a lifecycle that tracks each skill's downstream utility and automatically suppresses skills exhibiting persistently negative transfer.
\end{itemize}

\section{Related Work}
\label{sec:related}

\noindent\textbf{Workflow generation and optimization.}
A growing line of work constructs workflows dynamically at inference time \citep{yue2026static,yuksekgonul2025optimizing,khattab2023dspy,zhu2025survey}. Some methods search directly over workflow structure or optimize workflow graphs offline, from agent societies as optimizable graphs \citep{zhuge2024gptswarm} and meta-agent search \citep{hu2024adas} to MCTS-based optimization \citep{zhang2025aflow} and agentic supernets \citep{zhang2025multi}; recent systems add safety-constrained evolutionary programming \citep{zheng2025mermaidflow}, evolving multi-agent platforms \citep{wang2025evoagentx}, dynamic designer--executor workflows \citep{wang2026dyflow}, and deterministic routing orchestrators \citep{zhou2026orch}. We include these 4 systems, together with ADAS and AFlow, as baselines. The approaches above establish the importance of workflow structure but optimize for the current task or benchmark. We instead ask how successful workflows become reusable artifacts that remain available after the original task has finished.

\noindent\textbf{Memory for workflow agents.}
Another line augments agents with memories of prior plans, traces, or routines~\citep{zhang2025survey,xu2025mem}. Agent Workflow Memory~\citep{wang2024awm} is the closest in spirit, storing workflow-level experience to guide later generation, and \citet{sarukkai2026self} show that even naively accumulated self-generated trajectories improve sequential decision-making as in-context examples. Text-based memories improve planning, but they describe prior behavior rather than provide directly executable procedures. That gap motivates converting successful workflows into skill records that combine callable artifacts with structured guidance~\citep{wang2025reinforcement,xia2026skillrl}.

\noindent\textbf{Skills, tools, and executable reuse.}
A separate literature treats reusable procedures, tools, APIs, and skills as first-class objects that are more verifiable and reusable than text-only memories \citep{cai2023large, qian2023creator, yuan2023craft, wang2024trove, zheng2025skillweaver, ouyang2025code2mcp, di2026toolrosella}. However, these methods typically construct tools for a fixed task distribution \citep{yuan2023craft, wang2024trove} or decouple tool creation from the agent's own problem-solving trajectory \citep{qian2023creator, cai2023large,xia2026toolgenesis}. Recent skill-centric systems organize large pre-built corpora offline \citep{liang2026skillnet}, evolve skills across a multi-user population \citep{ma2026skillclaw}, or fine-tune weights alongside the library \citep{xia2026metaclaw}. FlowEvo differs on both axes. Its skills are compiled online from the agent's own verifier-passed traces, and the update is training-free.

\noindent\textbf{Training-free adaptation and self-evolution.}
Recent work explores how agents improve over time without weight updates, using search, self-generated feedback, memory revision, or evolving control state \citep{madaan2023self,putta2024agentq,yuan2024selfrewarding,zhang2026memrl,zhang2026liveevo,tian2024toward,yue2025don}. This line has early roots in late-2023--2024 systems that improve an agent's prompts, scaffolds, or collective topology at inference time. Promptbreeder~\citep{fernando2023promptbreeder} evolves prompt populations self-referentially, STOP~\citep{zelikman2024self} recursively rewrites the scaffolding program that invokes the model, and GPTSwarm~\citep{zhuge2024gptswarm} optimizes the communication graph of an agent society. In interactive environments, CLIN~\citep{majumder2024clin} maintains a continually revised memory of causal abstractions. Closest to our setting, Evoflux~\citep{bhandari2026evofluxinferencetimeevolutionexecutable} evolves executable tool workflows at inference time for compact agents. FlowEvo shares the training-free premise of this lineage, but the evolving object is more operationally grounded, namely an executable skill bank distilled from the agent's own successful workflows and governed by explicit admission and curation policies.

\noindent\textbf{Position of this work.}
Across these lines, one problem remains unresolved: how a workflow-based agent can persistently accumulate reusable capability from its own successful executions. Workflow optimization rarely preserves what was learned in executable form, text memory cannot be verified or invoked, tool creation happens outside the agent's own online loop, and self-evolution methods lack a concrete accumulation mechanism. FlowEvo addresses this gap by compiling successful workflows into reusable skill records and feeding them back through direct reuse and skill-conditioned workflow generation.

\section{Method}
\label{sec:method}

\subsection{Problem setup and the workflow-to-skill-to-workflow loop}
\label{subsec:problem_setup}

Consider a stream of tasks $\{x_t\}_{t=1}^{T}$. At episode $t$, the agent operates with an inference-time context $C_t = (B_t, M_t, R_t)$, where $B_t$ is the executable skill bank, $M_t$ denotes auxiliary memory layers, and $R_t$ collects the retrieval, routing, and failure state accumulated from prior use. Given task $x_t$, FlowEvo retrieves candidate artifacts, selects a routing mode, constructs a workflow plan $W_t$ when needed, executes and verifies the resulting program to obtain a trace $\tau_t$ and outcome $y_t$, and, when a successful episode yields reusable structure, compiles a candidate executable artifact and updates the memory state:
\begin{equation}
W_t = G(x_t; C_t), \;\; (\tau_t, y_t) = E(x_t, W_t), \;\; (B_{t+1}, M_{t+1}, R_{t+1}) = U(B_t, M_t, R_t, \tau_t, y_t).
\end{equation}
Self-evolution therefore occurs through updates to executable and non-executable memory, with the base model held fixed throughout the stream.

The central design choice is that successful workflows are compiled into executable skills rather than retained as text. A retrieved skill may be executed directly when compatibility and validation make reuse reliable enough, or exposed only as structured context guiding the generation of a new workflow. The skill bank is thus a persistent inference-time capability layer rather than prompt text alone. Algorithm~\ref{alg:flowevo} in Appendix~\ref{app:algorithm} presents the complete per-episode procedure in pseudocode with inline comments.

\noindent\textbf{System decomposition.}
Operationally, FlowEvo consists of 4 tightly coupled components: a workflow planner, an executor with verification, a trace-to-skill compiler, and a skill registry with lifecycle curation. The compiler extracts reusable executable artifacts only from verifier-passed successful episodes, while the registry maintains retrieval indices, admission state, and lifecycle updates. This decomposition makes the loop directly observable at the trace level and helps attribute downstream gains to specific stored skills rather than to untracked prompt history in the context window of the base model.

\subsection{Skill representation}
\label{subsec:representation}

A skill record in FlowEvo is a structured artifact that pairs an executable component with auxiliary textual guidance, rather than a free-form text note. The executable component is subject to interface, replay, and safety checks at admission where feasible, while the textual guidance supports skill-conditioned generation when direct execution is not selected. At a high level, each skill is $s = (f, \sigma, \mathcal{T}, m, \ell)$, where $f$ is the executable body with an identified entry point, $\sigma$ an explicit callable interface, $\mathcal{T}$ a set of replay or verifier tests, $m$ associated metadata, and $\ell$ the current lifecycle state of the record.

The metadata $m$ makes skill reuse explicit and auditable in practice. It records provenance (source task and fold), task-pattern tags, direct-reuse eligibility, preferred usage mode, reuse and fallback constraints, lightweight routing statistics, and negative evidence used to down-weight previously harmful skills. Not all fields are exposed in every route, but together they allow a retrieved skill to be called through a known interface, checked against stored validation data, and systematically tracked as its downstream behavior changes across later episodes of the stream that reuse or audit the skill.

The same stored artifact supports both reuse modes: direct execution relies on the interface, preconditions, eligibility for direct use, and replay evidence, while skill-conditioned generation exposes a compact structured view of the same artifact (signature, task pattern, scaffold hint, or a short code excerpt). The two uses have different reliability requirements and are therefore evaluated separately in the mechanism analyses.

\subsection{Workflow generation with retrieved skill context}
\label{subsec:planning}

For each task, a retriever scores candidate skills using lexical overlap, task-pattern compatibility, interface compatibility, and historical utility signals. The retriever is also aware of negative evidence, so previously harmful or mismatched skills are penalized rather than treated as fresh candidates at the next retrieval.

For clarity, we group the planner's behavior into 3 main routes: (i) dynamic generation with no retrieved skill, (ii) direct execution of a retrieved skill, and (iii) skill-conditioned generation in which retrieved skills are supplied only as structured context (Table~\ref{tab:planning_modes}, Appendix~\ref{app:algorithm}). Inside the third branch, the implementation contains finer-grained seeded modes. The distinction between routes (ii) and (iii) is central to the mechanism analysis.

Direct execution is permitted only for skills that remain compatible after filtering and validation. Skill-conditioned generation is used when retrieval is informative enough to guide planning but not reliable enough for direct execution, or when direct execution is intentionally disabled for analysis. The planner then injects a compact structured view of the skill rather than replaying the full artifact, which limits prompt overhead and makes gains attributable to structured reuse rather than unrestricted replay.

\subsection{Compiling successful workflows into executable skills}
\label{subsec:compile}

When a task succeeds, FlowEvo compiles the verified trace into a reusable executable artifact rather than a textual summary. The compiler identifies an entry point, infers a callable signature, extracts replay tests when possible, and records source and task-pattern tags together with metadata used for routing, such as reuse constraints and preferred usage mode. Compilation is attempted only from verifier-passed successful episodes, and the emitted candidate is passed to the admission layer, which may retain it for direct execution, retain it only for context exposure, or leave it in a shadow state for diagnostics. The result is a callable artifact that can be replayed, retrieved, and validated on later tasks in the stream.

\subsection{Admission and skill curation}
\label{subsec:lifecycle}

Not every successful trace should enter the active bank directly. Each candidate skill is first checked for duplicates and then passed through an admission layer governed by 3 diagnostic dimensions: \emph{interface compliance}, whether the extracted skill can be called under the intended signature and preconditions; \emph{functional correctness}, whether replay tests and held-out checks support the claimed behavior; and \emph{safety compliance}, whether the skill respects deployment constraints such as banned imports and banned calls (Appendix~\ref{app:constants}). Skills that fail these checks are not exposed for direct execution; borderline cases may be retained only for skill-conditioned use or shadowed for diagnosis.

Once admitted, skills remain subject to lifecycle control based on their \emph{downstream utility}: whether the skill improves later solving rather than merely existing in the bank. FlowEvo tracks utility, failure patterns, and audit outcomes, and can disable, restrict, repair, or prune a skill when repeated negative transfer accumulates. Current policies are conservative, re-checking suspicious skills with targeted audit episodes and downgrading direct-use privileges before removal. Unlike the admission dimensions, downstream utility cannot be evaluated at admission time and is assessed continuously after the skill enters the bank.

A key mechanism for detecting negative transfer is \emph{contrastive evaluation}: for each actively used skill, FlowEvo periodically compares the success rate of episodes where the skill was applied against matched episodes where it was withheld, and flags the skill for suppression when this contrastive delta is persistently negative. This provides a lightweight causal signal for identifying harmful skills without explicit holdout experiments or ground-truth labels for individual skill contributions. The signal complements causal failure attribution in broader multi-agent systems~\citep{ma2025automatic}.

\section{Experiments}
\label{sec:exp}

\begin{table}[t]
\centering
\small
\renewcommand{\arraystretch}{1.1}
\setlength{\tabcolsep}{2pt}
\begin{tabular*}{1\linewidth}{@{\extracolsep{\fill}}l rr rr rr rr rr@{}}
\toprule
        & \multicolumn{2}{c}{ALFWorld} & \multicolumn{2}{c}{HumanEval} & \multicolumn{2}{c}{MBPP} & \multicolumn{2}{c}{GSM8K} & \multicolumn{2}{c}{MATH-500} \\
\cmidrule(lr){2-3} \cmidrule(lr){4-5} \cmidrule(lr){6-7} \cmidrule(lr){8-9} \cmidrule(lr){10-11}
Method  & Acc. & Tok. & Acc. & Tok. & Acc. & Tok. & Acc. & Tok. & Acc. & Tok. \\
\midrule
ExpeL           & 46.3 & 32{,}958 & \underline{89.0} & \underline{883} & \underline{73.8} & \textbf{787} & \underline{92.1} & \underline{552} & \underline{67.5} & \textbf{1{,}260} \\
ADAS            & 53.0 & \underline{29{,}671} & 86.2 & 2{,}690 & 68.5 & \underline{2{,}228} & 90.7 & 2{,}255 & 58.2 & 2{,}733 \\
AFlow           & \underline{59.2} & 30{,}137 & 87.2 & 4{,}572 & 65.5 & 4{,}025 & 91.4 & 3{,}073 & 63.5 & 4{,}292 \\
\textbf{FlowEvo} & \textbf{85.6} & \textbf{9{,}329} & \textbf{95.1} & \textbf{880} & \textbf{79.6} & 2{,}230 & \textbf{97.1} & \textbf{541} & \textbf{75.9} & \underline{1{,}532} \\
\bottomrule
\end{tabular*}
\caption{Main results on the full standard splits of 5 benchmarks under a single shared GPT-4o-mini backbone. Accuracy is success rate (ALFWorld), pass@1 (HumanEval, MBPP), or solve rate (GSM8K, MATH-500); Tok.\ is average tokens per task. ALFWorld and MATH-500 FlowEvo entries are means over 3 seeds ($85.6 \pm 3.8$ and $75.9 \pm 0.3$). Bold marks the best and underline the second-best accuracy and token cost per benchmark.}
\label{tab:main_results}
\end{table}

We first compare FlowEvo with the 8 baselines on the full standard splits of the 5 benchmarks to demonstrate its accuracy and cost advantage (Section~\ref{subsec:main_results}). We then vary the backbone across 10 base models to show that the advantage does not depend on a single model (Section~\ref{subsec:robustness}). Finally, we isolate each mechanism through controlled ablations on ALFWorld, with further analyses of weak-supervision robustness, bank scalability, and accumulation dynamics (Sections~\ref{subsec:ablation} and~\ref{subsec:analysis}); complementary fixed-bank mechanism analyses on MBPP are reported in Appendix~\ref{app:mbpp} together with the protocol details.

\subsection{Experimental setup}
\label{subsec:setup}

\noindent\textbf{Benchmarks.}
We evaluate on the full standard splits of 5 benchmarks spanning 2 task families.
(1)~\emph{Interactive environment}: ALFWorld~\citep{shridhar2020alfworld}, a text-based household environment with 134 evaluation tasks from the \texttt{eval\_out\_of\_distribution} split covering 6 task types. The agent acts step by step (up to 50 steps) and receives only a binary terminal success signal, with no hidden tests, dense reward, or step-level annotation.
(2)~\emph{Code and math generation}: HumanEval~\citep{chen2021evaluating} (164 problems), MBPP~\citep{austin2021program} (standard 500-problem evaluation split), GSM8K~\citep{cobbe2021training} (1{,}319 test problems), and MATH-500~\citep{hendrycks2021measuring} (500 problems). Metrics are accuracy (success rate, pass@1, or solve rate) and average tokens per task.

\noindent\textbf{Baselines.}
We compare against 8 methods spanning 2 paradigms. Experience learning: ExpeL~\citep{zhao2024expel} and self-generated in-context examples (Self-Gen ICE)~\citep{sarukkai2026self}. Workflow optimization: ADAS~\citep{hu2024adas}, AFlow~\citep{zhang2025aflow}, MermaidFlow~\citep{zheng2025mermaidflow}, EvoAgentX~\citep{wang2025evoagentx}, DyFlow~\citep{wang2026dyflow}, and ORCH~\citep{zhou2026orch}. All methods share GPT-4o-mini as a single backbone; it is an inexpensive, widely used base and one of AFlow's own executor models, so a win under it cannot be dismissed as an artifact of a strong proprietary model, and Section~\ref{subsec:robustness} varies the backbone across 10 models. Every baseline runs from its official released code at default configuration under the same shared API endpoint, with matched optimization budgets (identical search and optimization iterations and per-task token caps; the full fidelity check is reported in Appendix~\ref{app:extended}).

\noindent\textbf{FlowEvo configuration and statistical reporting.}
FlowEvo evolves only through its inference-time capability layer, without any parameter updates to the backbone. Each benchmark runs independently from an empty initial bank, and no skills or memories transfer across benchmarks. On ALFWorld the agent accumulates skills online over the 134 tasks in sequence; on code and math benchmarks FlowEvo combines skill compilation with adaptive escalation, engaging more expensive strategies only when earlier stages fail verification. ALFWorld, MATH-500, the GPT-4.1 scale ladder, and all ablation and router analyses use 3 seeds, reported as means ($\pm$ sample std where shown), and paired accuracy comparisons are evaluated with McNemar's test.

\subsection{Main results}
\label{subsec:main_results}

\begin{table}[t]
\centering
\small
\renewcommand{\arraystretch}{1.1}
\begin{tabular*}{1\linewidth}{@{\extracolsep{\fill}}l cccc@{}}
\toprule
Method & HumanEval & MBPP & GSM8K & MATH-500 \\
\midrule
ExpeL~\citep{zhao2024expel}              & \underline{89.0} & \underline{73.8} & 92.1 & 67.5 \\
Self-Gen ICE~\citep{sarukkai2026self} & 86.2 & 66.1 & 81.2 & 67.3 \\
MermaidFlow~\citep{zheng2025mermaidflow} & 87.8 & 68.8 & 87.0 & 66.4 \\
EvoAgentX~\citep{wang2025evoagentx}      & 87.0 & 70.1 & 93.0 & \underline{73.6} \\
DyFlow~\citep{wang2026dyflow}            & 78.7 & 60.4 & 65.4 & 41.0 \\
ORCH~\citep{zhou2026orch}                & 87.8 & 70.8 & \underline{93.9} & 52.8 \\
ADAS~\citep{hu2024adas}                  & 86.2 & 68.5 & 90.7 & 58.2 \\
AFlow~\citep{zhang2025aflow}             & 87.2 & 65.5 & 91.4 & 63.5 \\
\textbf{FlowEvo}                  & \textbf{95.1} & \textbf{79.6} & \textbf{97.1} & \textbf{75.9} \\
\bottomrule
\end{tabular*}
\caption{Extended baseline panel on the 4 code/math full standard splits (accuracy as defined in Table~\ref{tab:main_results}; shared GPT-4o-mini backbone). Every baseline runs from its official released code at default configuration, so the lower scores of some recent methods reflect their default behavior under a deliberately modest backbone. Bold marks the best and underline the second best per column of the panel.}
\label{tab:recent_baselines}
\end{table}

Table~\ref{tab:main_results} presents the main comparison on the 5 full splits, and Table~\ref{tab:recent_baselines} extends the \mbox{code/math} comparison to recent workflow-optimization systems. FlowEvo attains the highest accuracy on all 5 benchmarks and the lowest token cost on 3. On MBPP~\citep{austin2021program} and MATH-500~\citep{hendrycks2021measuring} it trades noticeably more tokens per task than the text-memory baseline ExpeL~\citep{zhao2024expel} for accuracy gains of $+5.8$ and $+8.4$ points on the corresponding splits under the shared backbone.

\noindent\textbf{Interactive environments.}
On ALFWorld~\citep{shridhar2020alfworld}, FlowEvo attains 85.6\%, 26.4 points above AFlow~\citep{zhang2025aflow}. The margin arises because FlowEvo compiles successful action sequences into parameterized templates that replay directly on structurally similar tasks, while baselines re-derive solutions from scratch. Direct reuse also cuts average cost to 9{,}329 tokens per episode, roughly one third of any baseline's cost. ALFWorld offers no external verifier to fall back on, and executable compilation wins here by the widest margin observed across all 5 full standard splits.

\noindent\textbf{Code generation.}
On HumanEval~\citep{chen2021evaluating}, FlowEvo reaches 95.1\% at 880 tokens per task, 7.3 points above the strongest workflow-optimization baselines (MermaidFlow~\citep{zheng2025mermaidflow} and ORCH~\citep{zhou2026orch}, both 87.8\%) and 7.9 above AFlow~\citep{zhang2025aflow} at less than a fifth of its cost. The best baseline across both paradigms, ExpeL~\citep{zhao2024expel}, reaches 89.0\% at 883 tokens, 6.1 points below FlowEvo. AFlow~\citep{zhang2025aflow} and ADAS~\citep{hu2024adas} run fixed multi-stage pipelines of generation, testing, ensembling, review, and revision, costing several thousand tokens per task. FlowEvo's adaptive escalation instead solves most tasks in a single pass and engages more expensive strategies only on verification failure. On MBPP~\citep{austin2021program}, FlowEvo reaches 79.6\%, 5.8 points above the best baseline (ExpeL~\citep{zhao2024expel}, 73.8\%), at a cost comparable to ADAS~\citep{hu2024adas} and below that of AFlow~\citep{zhang2025aflow} on the same split, measured per task.

\noindent\textbf{Mathematical reasoning.}
On GSM8K~\citep{cobbe2021training}, FlowEvo reaches 97.1\% at 541 tokens per problem. ORCH~\citep{zhou2026orch} is closest at 93.9\%, followed by EvoAgentX~\citep{wang2025evoagentx} at 93.0\% and ExpeL~\citep{zhao2024expel} at 92.1\%. On MATH-500~\citep{hendrycks2021measuring}, FlowEvo reaches 75.9\% ($\pm 0.3$), 2.3 points above the strongest baseline (EvoAgentX~\citep{wang2025evoagentx}, 73.6\%) and 8.4 above ExpeL~\citep{zhao2024expel}, and leads ADAS~\citep{hu2024adas} and AFlow~\citep{zhang2025aflow} on all 7 subjects (Table~\ref{tab:math500}). Each problem on these benchmarks is effectively unique, so the direct-reuse route contributes little. The gain comes instead from adaptive escalation and from skill-conditioned generation supplying accumulated solution patterns as structured context.

\noindent\textbf{Performance across benchmark families.}
Each benchmark runs independently from an empty bank, so the comparison measures coverage across families rather than cross-domain transfer. Each baseline succeeds within its design regime and degrades outside it. ExpeL~\citep{zhao2024expel} is competitive on code and math but reaches 46.3\% on ALFWorld~\citep{shridhar2020alfworld}, where 6 task types recur with parametric variation that textual insights cannot replay. AFlow~\citep{zhang2025aflow} and ADAS~\citep{hu2024adas} reach 87.2\% and 86.2\% on HumanEval~\citep{chen2021evaluating} but 59.2\% and 53.0\% on ALFWorld~\citep{shridhar2020alfworld}, where a single optimized topology cannot cover 6 task types or adapt to step-by-step feedback. FlowEvo's two reuse modes cover both regimes. It Pareto-dominates ADAS~\citep{hu2024adas} and AFlow~\citep{zhang2025aflow}, the workflow-optimization baselines for which we also measure token cost, on 4 of 5 settings. The exception is MBPP~\citep{austin2021program}, where it gains accuracy at comparable rather than lower cost relative to those two baselines, as the token columns of Table~\ref{tab:main_results} show.

\subsection{Robustness across model scales and families}
\label{subsec:robustness}

\begin{table}[t]
\centering
\small
\renewcommand{\arraystretch}{1.1}
\setlength{\tabcolsep}{0.8pt}
\small
\begin{tabular*}{1\linewidth}{@{\extracolsep{\fill}}l ccccc@{}}
\toprule
Model & HumanEval & MBPP & GSM8K & MATH-500 & ALFWorld \\
\midrule
GPT-4.1-nano & 92.9$\pm$0.4\,(+2.6) & 71.1$\pm$1.1\,(+3.9) & 92.4$\pm$0.8\,(+8.5) & 69.8$\pm$1.0\,(+8.4) & 84.6$\pm$1.6\,(+53.2) \\
GPT-4.1-mini & 98.2$\pm$0.2\,(+4.9) & 77.9$\pm$0.5\,(+1.9) & 96.6$\pm$0.3\,(+6.8) & 74.8$\pm$0.6\,(+4.4) & 95.0$\pm$0.4\,(+29.4) \\
GPT-4.1      & 97.2$\pm$0.4\,(+3.9) & 82.1$\pm$0.2\,(+3.5) & 95.7$\pm$0.2\,(+2.5) & 74.5$\pm$0.5\,(+2.1) & 97.5$\pm$0.4\,(+21.3) \\
\midrule
Qwen2.5-7B-Instruct   & 87.2\,(+7.3)  & 69.0\,(+7.4) & 94.2\,(+3.3)  & 80.4\,(+11.4) & 84.3\,(+32.8) \\
Qwen3-8B              & 95.1\,(+1.8)  & 79.2\,(+2.8) & 98.4\,(+14.3) & 62.8\,($-$10.0) & 89.5\,(+20.2) \\
Llama-4-Scout         & 87.8\,(+12.2) & 70.2\,(+7.4) & 97.0\,(+1.9)  & 85.8\,(+15.4) & 88.1\,(+47.0) \\
Gemini-2.5-Flash-Lite & 95.7\,(+6.1)  & 76.0\,(+3.0) & 95.4\,(+11.2) & 60.8\,(+12.8) & 85.1\,(+28.4) \\
GPT-5.4-nano          & 95.1\,(+6.1)  & 73.2\,(+4.0) & 93.0\,(+8.1)  & 84.4\,(+7.8)  & 84.3\,(+35.1) \\
DeepSeek-V3.1         & 97.0\,(+3.1)  & 81.0\,(+2.8) & 98.3\,(+3.9)  & 78.6\,(+10.4) & 94.0\,(+10.5) \\
GPT-4o-mini           & 95.1\,(+6.1)  & 79.6\,(+5.8) & 97.1\,(+5.0)  & 75.9\,(+8.4)  & 85.6\,(+39.3) \\
\bottomrule
\end{tabular*}
\caption{FlowEvo across 10 base models, accuracy as defined in Table~\ref{tab:main_results}, with $\Delta$ over ExpeL in parentheses. Top block: the GPT-4.1 scale ladder (same family and generation, only size varies), mean $\pm$ sample std over 3 seeds. Bottom block: 7 further models spanning 5 families and scales from 7B to 671B parameters, dense to mixture-of-experts.}
\label{tab:backbones}
\end{table}

\noindent\textbf{Across model scales.}
Skill compilation and admission depend on the base model's ability to produce clean, generalizable code and follow interface conventions, so we test a controlled scale ladder and a breadth panel across families (Table~\ref{tab:backbones}). The GPT-4.1 ladder varies scale while holding family and generation fixed. On the 3 datasets that exercise the full skill loop (GSM8K~\citep{cobbe2021training}, MATH-500~\citep{hendrycks2021measuring}, and ALFWorld~\citep{shridhar2020alfworld}), the gain over ExpeL~\citep{zhao2024expel} \emph{grows} as model size decreases, most sharply on ALFWorld, where the smallest model gains $+53.2$ points. The gains therefore do not require a strong base model. On HumanEval~\citep{chen2021evaluating} and MBPP~\citep{austin2021program} all 3 sizes sit near saturation, so the remaining differences are correspondingly small.

\noindent\textbf{Across model families.}
The remaining 7 models span 5 families and 7B to 671B parameters, dense to mixture-of-experts. Over the full panel of 10 models (50 paired model--dataset cells), FlowEvo beats ExpeL~\citep{zhao2024expel} in 49, with mean $\Delta$ $+11.0$ and median $+7.0$ points. The single cell below baseline is Qwen3-8B on MATH-500~\citep{hendrycks2021measuring} ($-10.0$), a small model on hard reasoning, where interface-violating skill bodies are rejected at admission, leaving the MATH bank too sparse to provide usable skills for the problems that follow in the remaining evaluation stream.

\subsection{Ablation study: 3 mechanisms}
\label{subsec:ablation}

\begin{table}[t]
\centering
\small
\renewcommand{\arraystretch}{1.1}
\begin{tabular*}{1\linewidth}{@{\extracolsep{\fill}}l l c c@{}}
\toprule
Stage & Mechanism added & Success rate (\%) & $\Delta$ \\
\midrule
1 & ReAct (baseline, no skill bank)                      & 33.6 & --- \\
2 & $+$ workflow-to-skill compilation (storage, no reuse) & 38.8 & $+5.2$ \\
3 & $+$ skill-to-workflow feedback (direct $+$ conditioned) & 80.6 & $+41.8$ \\
4 & $+$ skill curation (lifecycle management)             & 85.6 & $+5.0$ \\
\bottomrule
\end{tabular*}
\caption{Per-component ablation on ALFWorld (134 tasks, 3 seeds). Starting from ReAct, one mechanism is switched on at a time, so each row's increment is attributable to the mechanism added at that stage of the sequence.}
\label{tab:ablation}
\end{table}

Table~\ref{tab:ablation} isolates each mechanism through controlled ablations against ReAct~\citep{yao2022react}, which uses interleaved reasoning and acting without skill accumulation. Switching on one mechanism at a time contributes $+5.2$ points for compilation, $+41.8$ for skill-to-workflow feedback, and $+5.0$ for curation. Compilation without reuse already helps because extracting reusable structure from successful trajectories regularizes behavior, improving auxiliary memories and intermediate representations.

\noindent\textbf{Skill-to-workflow feedback.}
Enabling skill reuse on top of compilation produces the dominant improvement (38.8\% to 80.6\%), confirming that the gain is driven by skill feedback rather than curation. Compiled skills are retrieved and either executed directly or injected as structured context. On ALFWorld, where task types recur with parametric variation, direct reuse dominates the routing mix (Section~\ref{subsec:analysis}).

\noindent\textbf{Skill curation.}
Curation contributes $+5.0$ points and a qualitatively important safeguard against negative transfer. In one seed of the 3-seed campaign, contrastive evaluation identified the \texttt{pick\_two\_obj\_and\_place} skill as harmful: guided success was 2/14 (14\%) versus 1/3 (33\%) unguided, a delta of $-0.19$, below the suppression threshold of $-0.1$, triggering suppression within 14 guided episodes (Appendices~\ref{app:extended} and~\ref{app:routing}). Without curation this skill would keep harming subsequent episodes. With curation, routing for this task type defaults to dynamic generation. Skill accumulation is thus not uniformly beneficial, and curation keeps the bank from amplifying its own worst entries.

\noindent\textbf{Robustness to weak or noisy supervision.}
Compilation is gated on verifier-passed traces, so one may worry that FlowEvo requires clean success signals. A 3-condition ablation of the admission gate (3 seeds; Table~\ref{tab:noise}) shows otherwise: removing the trace-success requirement costs $1.5$ points (85.6\% $\to$ 84.1\%), and flipping 20\% of pass/fail labels in both directions costs $1.7$ points. Neither change is significant under McNemar's test ($p > 0.1$). The 2-stage design, which pairs a verifier pass at compile time with contrastive checking during use, absorbs upstream label noise without extra supervision.

\noindent\textbf{Skill-bank scalability.}
Across the 134 episodes the bank saturates at 7 workflow templates (one per task type plus one blueprint entry) and 18 exemplars. The first 28 episodes cover 5 task types, and the rest add a single entry (Tables~\ref{tab:growth} and~\ref{tab:curation}). De-duplication and contrastive suppression keep the bank compact and retrieval cost bounded.

\subsection{Analysis of skill accumulation dynamics}
\label{subsec:analysis}

\begin{figure}[t]
\centering
\begin{minipage}[c]{0.48\linewidth}
    \centering
    \includegraphics[width=\linewidth]{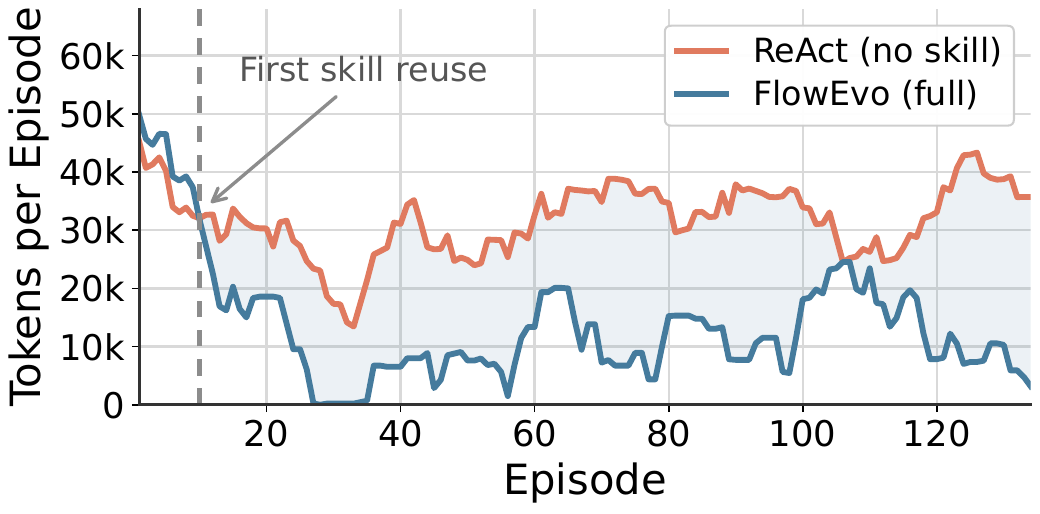}\\
    \small (a) Token cost per episode
\end{minipage}
\hfill
\begin{minipage}[c]{0.48\linewidth}
    \centering
    \includegraphics[width=\linewidth]{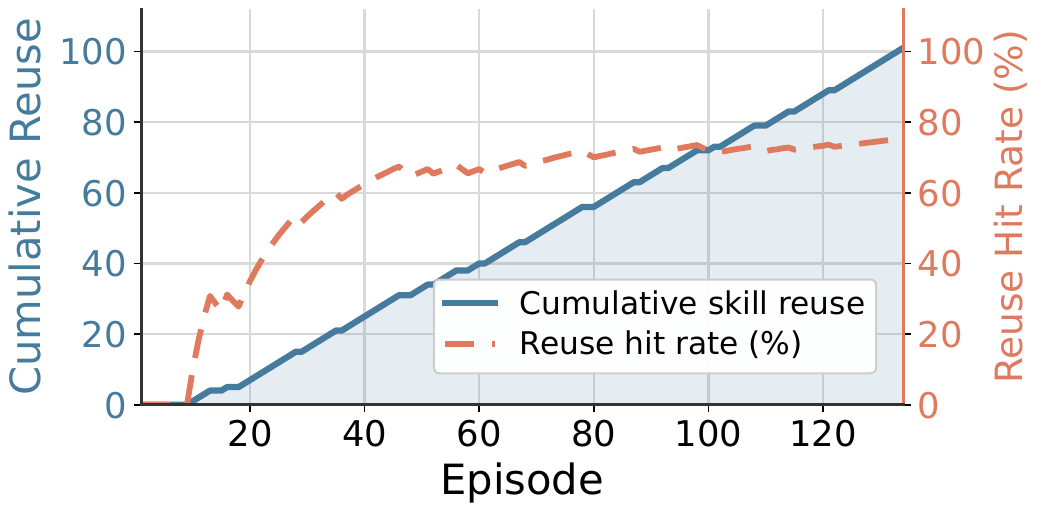}
    \small (b) Cumulative reuse and hit rate
\end{minipage}
\caption{Skill accumulation dynamics on ALFWorld (representative run). (a)~Token cost drops sharply once the first skill reuse occurs around episode~10 (dashed line) and stays at or below ReAct. (b)~Cumulative direct-reuse count (blue, solid) grows nearly linearly to 101/134; the reuse hit rate (orange, dashed) stabilizes around 75\%.}
\label{fig:dynamics}
\end{figure}

\noindent\textbf{Token efficiency and reuse.}
Figure~\ref{fig:dynamics} plots the representative run of Appendices~\ref{app:skill-record}--\ref{app:routing}. After the first skill reuse occurs (around episode~10), token consumption drops sharply and stays low, because direct reuse bypasses exploratory interaction. Averaged over 3 seeds, FlowEvo uses 9{,}329 tokens per episode against 29{,}671--32{,}958 for the baselines. The cumulative direct-reuse count grows steadily to 101 of 134 episodes, the hit rate stabilizes around 75\%, and 99 of the 101 direct-reuse episodes succeed (98\%).

\noindent\textbf{Per-task-type improvement.}
Decomposing the success rate by task type in Figure~\ref{fig:per_task} and Appendix~\ref{app:extended} shows that the improvement tracks the direct-hit rate, that is, how often the compiled skill completed the task through direct execution alone. Gains fall from $+78$ points at a direct-hit rate of 83\% on \texttt{pick\_heat\_then\_place\_in\_recep} to $+28$ points at 0\% on \texttt{look\_at\_obj\_in\_light}. The latter isolates the second reuse mode: the skill matched on 15 of 18 episodes but completed none through direct execution (Tables~\ref{tab:skill-bank}--\ref{tab:routing}). The sixth type, \texttt{pick\_two\_obj\_and\_place}, was suppressed by curation (Section~\ref{subsec:ablation}).

\noindent\textbf{Fixed-bank mechanism analyses on MBPP.}
Appendix~\ref{app:mbpp} isolates the two reuse modes under a growth-to-frozen MBPP protocol. On exact-transfer cases, executable reuse beats dynamic generation on both accuracy and cost (85.9\% at 204 tokens vs.\ 81.7\% at 224), while text workflow memory trades $+1.4$ points for 85\% more tokens. On the full evaluation split, no router pair separates under paired testing (all pairwise McNemar $p > 0.5$).

\section{Discussion}
\label{sec:discussion}

Accumulated skills help through 2 complementary mechanisms. The dominant one is direct skill reuse. On ALFWorld~\citep{shridhar2020alfworld}, compiled skills are replayed on the majority of episodes at a 98\% success rate in the representative run, consistent with the $+41.8$-point skill-feedback gain over 3 seeds. The second is skill-conditioned generation when direct replay is unavailable. This mode is the relevant one on code and math, isolated by the \texttt{look\_at\_obj\_in\_light} case and the fixed-bank MBPP~\citep{austin2021program} analyses (Appendix~\ref{app:mbpp}), which freeze the bank before evaluation.

FlowEvo also clarifies the relation between workflow optimization and skill learning, preserving the stable parts of a successful workflow as a persistent executable object with an interface, validation tests, and usage history. The comparison with text memory is not one-sided: on the MBPP~\citep{austin2021program} exact-transfer slice, text workflow memory attains a slightly higher pass rate at much higher cost. At comparable accuracy, however, executable skills give a better accuracy--efficiency frontier, together with verifiability, invocability, and auditability that a text record, however detailed, cannot provide.

The backbone sweep indicates where an inference-time capability layer matters most. Across 10 base models the gain over ExpeL~\citep{zhao2024expel} is positive in 49 of 50 model--dataset cells, and within the GPT-4.1 ladder it widens as model size decreases, reaching $+53.2$ points on ALFWorld~\citep{shridhar2020alfworld} for the smallest model while HumanEval~\citep{chen2021evaluating} and MBPP~\citep{austin2021program} sit near saturation. The ladder holds family and generation fixed, so the widening gap tracks scale rather than a family idiosyncrasy, and the breadth panel repeats the direction across dense and mixture-of-experts designs, with a mean gain of $+11.0$ points and a median of $+7.0$ over the full 50 cells. Median and mean agree in sign, so the improvement is broad rather than concentrated in a few cells.

The pattern is consistent with compiled skills supplying structure that a stronger backbone can reconstruct on demand. The largest gains therefore fall where the base model is weakest, which is also where inference is least expensive. The single trailing cell, Qwen3-8B on MATH-500~\citep{hendrycks2021measuring}, marks the lower end of that range, where too few candidate skills survive admission to populate the bank and support retrieval. Above that floor the loop engages without further requirements on the backbone, and the advantage narrows only where the base model alone already sits near ceiling accuracy, as on HumanEval~\citep{chen2021evaluating} and MBPP~\citep{austin2021program}. The capability layer is thus complementary to scale: it substitutes structure where the backbone lacks it and steps aside where the backbone can already reconstruct that structure on demand.

The current evidence should guide interpretation. Our experiments rely on domains where correctness can be checked with environment feedback or hidden tests. The ALFWorld results and the admission-gate ablation show that the loop tolerates weak and even noisy signals. Fully signal-free settings such as free-form dialogue and long-form writing remain out of scope, and delayed, partial, or adversarial feedback may pose further challenges~\citep{wang2026ai}. The framework also needs enough base capability to bootstrap the loop (see the Qwen3-8B MATH-500 cell in Table~\ref{tab:backbones}). There are 2 directions. First, self-evolution could extend beyond task-solving skills to the rest of the agent harness~\citep{guo2026drevo}, including verification helpers, retry policies, and retrieval heuristics, and to broader tool ecosystems via standardized protocols~\citep{ouyang2025code2mcp,di2026toolrosella}. Second, bank maintenance under long-running use is a problem in its own right, requiring skills to be revised, merged, and retired while limiting retrieval interference~\citep{osti_3366326,guo2025dior}.

\section{Conclusion}
\label{sec:conclusion}

FlowEvo compiles successful workflows into reusable skill records for direct execution or structured context. Across 5 full-split benchmarks, it achieves the highest accuracy among 8 baselines and Pareto-dominates ADAS~\citep{hu2024adas} and AFlow~\citep{zhang2025aflow} on 4 of 5. Gains persist across 10 base models; ablations identify skill feedback as the main contributor. Agents can improve at inference time by retaining reusable execution structure.

\newpage

\bibliographystyle{colm2026_conference}
\bibliography{main}

\clearpage
\appendix
\renewcommand{\topfraction}{0.9}
\renewcommand{\bottomfraction}{0.7}
\renewcommand{\floatpagefraction}{0.75}
\renewcommand{\textfraction}{0.1}
\raggedbottom

\section{Fixed-Bank Mechanism Analyses on MBPP}
\label{app:mbpp}

The main-text ALFWorld analyses study skill accumulation \emph{online}, as the bank grows. This appendix isolates the two reuse modes under a frozen bank, so that downstream effects are attributable to previously accumulated capabilities rather than to continued adaptation.

\noindent\textbf{Fixed-bank protocol.}
The MBPP mechanism analyses use a deterministic 3-fold growth-to-frozen protocol over a fixed task universe. In each fold, one partition is used during a growth phase to accumulate skills and workflow memories, while the held-out partition is reserved for fixed-bank evaluation. During growth, FlowEvo may compile new skills, summarize workflows, and revise routing statistics. During the subsequent fixed-bank phase, all such updates are disabled: no new skills are compiled, no routing statistics are revised, and no additional workflow memories are added.

We study 2 held-out settings on this MBPP universe. \emph{Exact-transfer} evaluation aggregates 71 held-out cases that admit direct execution of a compatible previously acquired skill. \emph{Compatible non-exact} evaluation uses 44 held-out tasks that do not admit exact replay but do admit transfer from a related acquired skill through structured context; each such task is evaluated over 3 fresh task orders, yielding 132 fixed-bank episodes in total.

\subsection{Exact-transfer analysis: executable skills as efficient direct reuse}
\label{app:mbpp-exact}

Table~\ref{tab:exact_direct} evaluates held-out cases in which a compatible previously acquired skill can be executed directly. Under this protocol, FlowEvo selects the direct-execution route on all 71 cases. Relative to dynamic generation, executable reuse improves pass rate from 81.7\% to 85.9\% while reducing average token usage from 224.2 to 204.4. Text workflow memory attains a slightly higher pass rate at 87.3\%, but does so at substantially greater cost: it leads executable reuse by 1.4 points while using 85\% more tokens.

\begin{table}[!htbp]
    \centering
    \small
    \renewcommand{\arraystretch}{1.1}
    \begin{tabular*}{0.72\linewidth}{@{\extracolsep{\fill}}lcc@{}}
        \toprule
        Method & Pass rate (\%) & Avg.\ tokens \\
        \midrule
        Dynamic generation     & 81.7 & 224.2 \\
        Text workflow memory   & 87.3 & 377.5 \\
        Executable skill reuse & 85.9 & 204.4 \\
        \bottomrule
    \end{tabular*}
    \caption{MBPP exact-transfer evaluation. Each example admits direct reuse of a compatible previously acquired skill. Scores are computed during the fixed-bank phase on hidden tests.}
    \label{tab:exact_direct}
\end{table}

The conclusion is therefore not that executable reuse uniformly dominates every comparison point. It is that the two memory forms sit on an accuracy--efficiency frontier, and at comparable accuracy, executable skills give the stronger frontier: they improve on dynamic generation in both accuracy and efficiency, remain markedly more efficient than text-based workflow memory at similar task success, and add verifiability, invocability, and auditability that a text record, however detailed, cannot provide.

\subsection{Compatible non-exact analysis: skills as structured context}
\label{app:mbpp-nonexact}

Table~\ref{tab:compatible_non_exact} studies the more difficult setting in which exact replay is unavailable, so any effect must arise from how retrieved skills condition the workflow constructor, and Table~\ref{tab:priority_conflict} sharpens the route-choice question on episodes where direct-first and context-first priorities disagree most clearly after retrieval. Unlike the main non-exact table, which is defined by a task-level compatibility filter, the priority-conflict diagnostic is formed by episode-level regrouping under a matched fixed-bank evaluation budget; the total remains 132 episodes because the comparison is budget-matched at the episode level.

\begin{table}[!htbp]
    \centering
    \small
    \renewcommand{\arraystretch}{1.1}
    \setlength{\tabcolsep}{6pt}
    \begin{tabular}{>{\raggedright\arraybackslash}p{0.48\linewidth}ccc}
        \toprule
        Method & First pass (\%) & Avg.\ retries & Avg.\ tokens \\
        \midrule
        FlowEvo (direct-preferred router) & 75.8 & 0.159 & 215.0 \\
        FlowEvo (seed-only router)        & 75.8 & 0.197 & 233.4 \\
        FlowEvo (bifurcated router)       & 78.0 & 0.182 & 224.8 \\
        \bottomrule
    \end{tabular}
    \caption{MBPP compatible non-exact evaluation. These tasks do not admit exact direct replay, but can benefit from related previously acquired skills exposed as structured context.}
    \label{tab:compatible_non_exact}
\end{table}

\begin{table}[!htbp]
\centering
\small
\begin{tabularx}{\linewidth}{@{}Xccc@{}}
\toprule
Method & First pass (\%) & Avg.\ retries & Avg.\ tokens \\
\midrule
FlowEvo (direct-preferred router) & 75.0 & 0.136 & 213.8 \\
FlowEvo (seed-only router) & 78.0 & 0.159 & 224.9  \\
FlowEvo (bifurcated router) & 75.8 & 0.182 & 221.2  \\
FlowEvo (exact-or-seeded router) & 75.0 & 0.205 & 230.9  \\
\bottomrule
\end{tabularx}
\caption{MBPP priority-conflict slice under the same fixed-bank episode budget as Table~\ref{tab:compatible_non_exact}. Regrouping is episode-level, so it is not a literal task subset.}
\label{tab:priority_conflict}
\end{table}

The absolute differences across router variants are small, and paired testing makes their status explicit. Because the fixed-bank slices above are small (132 episodes), we run the paired tests on the full MBPP evaluation split, where all 4 routers can be compared on identical items over 3 seeds (Table~\ref{tab:mcnemar}); the per-pair differences there are correspondingly smaller than the raw gaps in Tables~\ref{tab:compatible_non_exact}--\ref{tab:priority_conflict}. On that basis, all 6 router-pair comparisons have an absolute difference below 0.6 points, a 95\% confidence interval that crosses zero, and McNemar $p$ above 0.5. These near-identical numbers are the intended finding: skill-as-context is \emph{robust to router choice} in the no-direct-replay regime, so routing policy is second-order and does not change the main conclusions. The router effect that does matter appears in the online ALFWorld setting instead, where the direct-hit spectrum of Section~\ref{subsec:analysis} shows the skill bank improving accuracy by $+28$ points on a task type it never completes through direct execution (\texttt{look\_at\_obj\_in\_light}, Table~\ref{tab:routing}).

\begin{table}[!htbp]
\centering
\small
\renewcommand{\arraystretch}{1.1}
\begin{tabular*}{1\linewidth}{@{\extracolsep{\fill}}l c c c@{}}
\toprule
Pair & $\Delta$ first-pass & 95\% CI & McNemar $p$ \\
\midrule
direct-preferred vs.\ seed-only        & $-0.35$ & $[-1.58, +0.88]$ & 0.789 \\
direct-preferred vs.\ bifurcated       & $-0.18$ & $[-1.05, +0.70]$ & 1.000 \\
direct-preferred vs.\ exact-or-seeded  & $+0.18$ & $[-0.70, +1.05]$ & 1.000 \\
seed-only vs.\ bifurcated              & $+0.18$ & $[-1.23, +1.76]$ & 1.000 \\
seed-only vs.\ exact-or-seeded         & $+0.53$ & $[-0.88, +1.93]$ & 0.628 \\
bifurcated vs.\ exact-or-seeded        & $+0.35$ & $[-0.88, +1.58]$ & 0.773 \\
\bottomrule
\end{tabular*}
\caption{Within-family router-pair statistics on the MBPP evaluation split over 3 seeds. $\Delta$ and the 95\% confidence interval are in percentage points.}
\label{tab:mcnemar}
\end{table}

\section{Extended Results and Robustness Details}
\label{app:extended}

\subsection{Baseline fidelity and MATH-500 per-subject breakdown}
\label{app:math500}

AFlow and ADAS use their original paper configurations, with the same search rounds and operator sets, under matched optimization budgets (identical search/optimization iterations and per-task token caps). As a fidelity check on our baseline reruns, the ADAS rerun on GSM8K reproduces the number reported in the ADAS paper (90.7\% versus 90.8\%).

\begin{table}[!htbp]
\centering
\scriptsize
\renewcommand{\arraystretch}{1.1}
\setlength{\tabcolsep}{2.2pt}
\begin{tabular*}{1\linewidth}{@{\extracolsep{\fill}}l cccccccc@{}}
\toprule
Method & Algebra & Inter.\,Alg. & Prealgebra & Count\,\&\,Prob & Geometry & Num.\,Theory & Precalc. & Overall \\
\midrule
ADAS    & 74.19\,$\pm$\,1.40 & 35.05\,$\pm$\,2.73 & 71.54\,$\pm$\,1.41 & 62.28\,$\pm$\,1.52 & \underline{52.03}\,$\pm$\,1.41 & 66.13\,$\pm$\,5.82 & 36.31\,$\pm$\,5.46 & 58.20\,$\pm$\,0.20 \\
AFlow   & \underline{76.08}\,$\pm$\,1.68 & \underline{46.39}\,$\pm$\,1.79 & \underline{75.20}\,$\pm$\,0.70 & \underline{71.05}\,$\pm$\,2.63 & 45.53\,$\pm$\,1.41 & \underline{80.11}\,$\pm$\,2.46 & \underline{37.50}\,$\pm$\,1.79 & \underline{63.47}\,$\pm$\,1.10 \\
\textbf{FlowEvo} & \textbf{93.01}\,$\pm$\,0.93 & \textbf{53.61}\,$\pm$\,3.72 & \textbf{84.96}\,$\pm$\,1.86 & \textbf{73.68}\,$\pm$\,2.63 & \textbf{65.85}\,$\pm$\,6.45 & \textbf{88.71}\,$\pm$\,1.61 & \textbf{58.33}\,$\pm$\,2.06 & \textbf{75.93}\,$\pm$\,0.31 \\
\bottomrule
\end{tabular*}
\caption{MATH-500 ($n=500$) per-subject accuracy (\%), mean $\pm$ sample std over 3 seeds, GPT-4o-mini backbone. FlowEvo leads on every subject, so the overall gap is not driven by a single subject. The per-subject splits are small (e.g., Geometry $n=41$, Counting\,\&\,Probability $n=38$, Precalculus $n=56$), so their standard deviations run higher than the overall figure's. Bold marks the best and underline the second best per column.}
\label{tab:math500}
\end{table}

\subsection{Admission-gate ablation under label noise}
\label{app:noise}

The 2-stage admission (verifier-pass at compile time, contrastive check during use) absorbs upstream label noise: harmful admissions that slip through a corrupted gate are caught downstream by contrastive curation in later episodes.

\begin{table}[!htbp]
\centering
\small
\renewcommand{\arraystretch}{1.1}
\begin{tabular*}{1\linewidth}{@{\extracolsep{\fill}}l c c c@{}}
\toprule
Condition & Success rate & $\Delta$ from full library & McNemar $p$ \\
\midrule
Full library (anchor)                          & 85.6\,$\pm$\,3.8\% & ---     & --- \\
No trace-success requirement                   & 84.1\,$\pm$\,1.9\% & $-1.5$\% & 0.125 \\
20\% false-positive/false-negative label flip  & 83.8\,$\pm$\,2.2\% & $-1.7$\% & 0.227 \\
\bottomrule
\end{tabular*}
\caption{Admission-gate ablation on ALFWorld (3 seeds). Removing the trace-success requirement, or flipping 20\% of the pass/fail labels in both directions, costs only 1.5--1.7 points and is not statistically significant under McNemar's test. Differences are computed at unrounded precision, so they can deviate from the rounded column values by 0.1.}
\label{tab:noise}
\end{table}

\subsection{Skill-bank growth and curation decisions}
\label{app:growth}

\begin{table}[!htbp]
\centering
\small
\renewcommand{\arraystretch}{1.1}
\begin{tabular*}{1\linewidth}{@{\extracolsep{\fill}}c l c c@{}}
\toprule
Episode & Triggering task type & Workflows learned & Exemplars total \\
\midrule
11  & \texttt{pick\_and\_place\_simple}            & $0 \to 1$ & 11 \\
15  & \texttt{pick\_clean\_then\_place\_in\_recep} & $1 \to 2$ & 12 \\
19  & \texttt{pick\_heat\_then\_place\_in\_recep}  & $2 \to 3$ & 14 \\
23  & \texttt{pick\_cool\_then\_place\_in\_recep}  & $3 \to 4$ & 15 \\
28  & \texttt{look\_at\_obj\_in\_light}            & $4 \to 5$ & 16 \\
101 & \texttt{pick\_two\_obj\_and\_place}          & $5 \to 6$ & 17 \\
134 (final) & ---                                  & 6 learned $+$ 1 blueprint & 18 \\
\bottomrule
\end{tabular*}
\caption{Skill-bank growth across 134 ALFWorld episodes, from one seed of the 3-seed campaign (the same run as Table~\ref{tab:curation}). The bank saturates at 7 workflow templates (6 task-type templates plus one blueprint entry) and 18 exemplars; the first 28 episodes cover 5 task types, and the remaining episodes add one further template.}
\label{tab:growth}
\end{table}

\begin{table}[!htbp]
\centering
\small
\renewcommand{\arraystretch}{1.1}
\setlength{\tabcolsep}{3pt}
\begin{tabular*}{1\linewidth}{@{\extracolsep{\fill}}l c c c l@{}}
\toprule
Task type & Guided & Holdout & $\Delta$ & Curation decision \\
\midrule
\texttt{look\_at\_obj\_in\_light}            & 17/18 (94\%)  & 0/0        & ---     & retain (no holdout) \\
\texttt{pick\_and\_place\_simple}            & 20/20 (100\%) & 4/4 (100\%) & 0\%     & neutral, retain \\
\texttt{pick\_clean\_then\_place\_in\_recep} & 26/26 (100\%) & 5/5 (100\%) & 0\%     & neutral, retain \\
\texttt{pick\_cool\_then\_place\_in\_recep}  & 16/18 (89\%)  & 3/3 (100\%) & $-11$\% & matched $\Delta$ above gate \\
\texttt{pick\_heat\_then\_place\_in\_recep}  & 18/19 (95\%)  & 4/4 (100\%) & $-5$\%  & no trigger \\
\texttt{pick\_two\_obj\_and\_place}          & 2/14 (14\%)   & 1/3 (33\%)  & $-19$\% & suppression triggered \\
\bottomrule
\end{tabular*}
\caption{Per-task-type contrastive curation decisions, from the same campaign seed as Table~\ref{tab:growth}; the representative run of Tables~\ref{tab:skill-bank} and~\ref{tab:routing} is a separate run, so per-type success counts differ between the two. Guided and Holdout are cumulative outcomes over all episodes of the task type, and $\Delta$ is their difference; \texttt{look\_at\_obj\_in\_light} has no withheld episodes, so no contrastive delta is defined for it. The suppression trigger itself is evaluated on the matched subset of guided versus withheld episodes maintained per skill (\texttt{contrastive\_stats}, Appendix~\ref{app:skill-record}): for \texttt{pick\_two\_obj\_and\_place} that subset stood at 2/14 guided versus 1/3 withheld ($\Delta = -0.19$) when the trigger fired, and guided counts stop accruing after suppression, so its cumulative row equals the at-trigger values, while a negative cumulative $\Delta$ elsewhere does not by itself imply suppression. The final bank keeps the suppressed template as a shadow entry for diagnostics, out of active retrieval.}
\label{tab:curation}
\end{table}

\clearpage
\subsection{Per-task-type improvement (representative run)}
\label{app:pertask}

\begin{wrapfigure}{r}{0.48\linewidth}
\centering
\includegraphics[width=\linewidth]{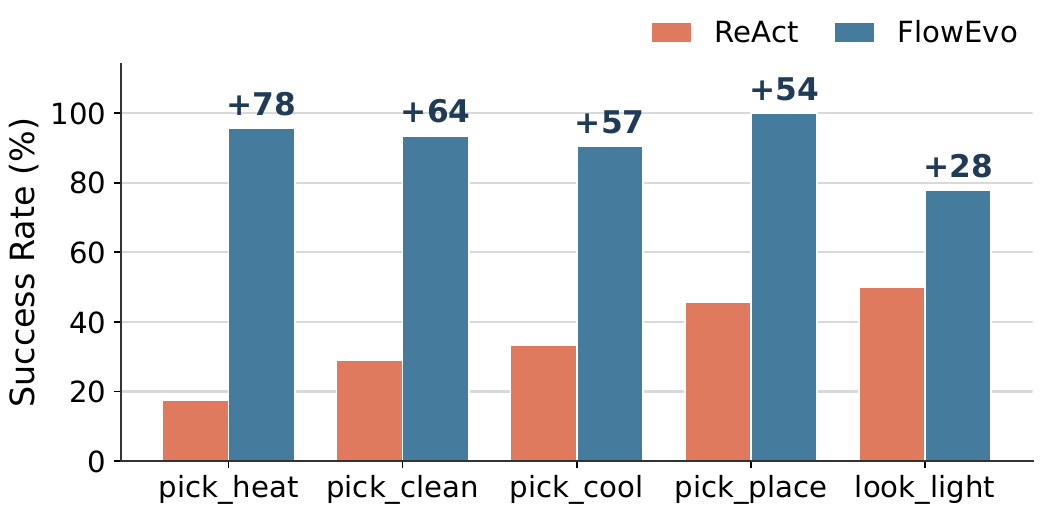}
\caption{Per-task-type success rate on ALFWorld (GPT-4o-mini; representative run; $n$ per type varies).}
\label{fig:per_task}
\end{wrapfigure}
Figure~\ref{fig:per_task} gives the per-task-type decomposition summarized in Section~\ref{subsec:analysis}. FlowEvo improves over ReAct on all 5 displayed task types, with gains ranging from $+78$ points on \texttt{pick\_heat\_then\_place\_in\_recep} (83\% direct-hit rate) to $+28$ on \texttt{look\_at\_obj\_in\_light} (0\% direct-hit rate), tracing an approximately monotone relation between how often a compiled skill executes directly and how much it helps; the routing counts behind each direct-hit rate are in Table~\ref{tab:routing}. The \texttt{pick\_two} task type is omitted from the figure because its compiled skill was suppressed by skill curation as a case of negative transfer during the run, as discussed in Section~\ref{subsec:ablation}. Axis labels abbreviate the task-type names, which are listed in full in Tables~\ref{tab:skill-bank}--\ref{tab:routing}.

\section{Per-Episode Algorithm}
\label{app:algorithm}

Algorithm~\ref{alg:flowevo} gives the per-episode procedure referenced in Section~\ref{subsec:problem_setup}, showing how retrieval, routing, compilation, admission, and skill curation are ordered within a single pass of the inference-time loop for one incoming task.

\begin{algorithm}[!htbp]
\caption{FlowEvo workflow--skill--workflow loop at inference time.}
\label{alg:flowevo}
\begin{algorithmic}[1]
\Require Task stream $\{x_t\}_{t=1}^{T}$; a fixed base model
\State Initialize skill bank $B_1 \gets \emptyset$; auxiliary memory $M_1 \gets \emptyset$; routing state $R_1 \gets \emptyset$
\For{$t = 1, \ldots, T$}
    \State $\mathcal{S}_t \gets \textsc{Retrieve}(x_t, B_t, R_t)$ \Comment{score candidates by pattern, interface, and utility}
    \State $r_t \gets \textsc{RouteSelect}(x_t, \mathcal{S}_t)$ \Comment{dynamic / direct-execute / skill-conditioned}
    \If{$r_t = $ direct-execute \textbf{and} $\textsc{Compatible}(x_t, \mathcal{S}_t)$}
        \State $W_t \gets$ call retrieved skill $s^\star \in \mathcal{S}_t$ as an executable subroutine
    \ElsIf{$r_t \neq $ dynamic}
        \State $W_t \gets G(x_t;\, C_t = (B_t, M_t, R_t))$ \Comment{generate with structured skill context}
    \Else
        \State $W_t \gets G(x_t;\, \varnothing)$ \Comment{dynamic generation without skill context}
    \EndIf
    \State $(\tau_t, y_t) \gets E(x_t, W_t)$ \Comment{execute and verify; $y_t$ indicates success}
    \If{$y_t = \text{success}$}
        \State $\hat{s}_t \gets \textsc{Compile}(\tau_t)$ \Comment{extract entry point, interface, replay tests, metadata}
        \State $B_{t+1} \gets \textsc{Admit}(B_t, \hat{s}_t)$ \Comment{duplicate check; interface, replay, safety validation}
    \Else
        \State $B_{t+1} \gets B_t$
    \EndIf
    \State $M_{t+1}, R_{t+1} \gets \textsc{Update}(M_t, R_t, \tau_t, y_t, \mathcal{S}_t)$ \Comment{update usage stats, negative evidence}
    \State $B_{t+1} \gets \textsc{Curate}(B_{t+1}, R_{t+1})$ \Comment{suppress persistently harmful skills}
\EndFor
\State \textbf{return} final skill bank $B_{T+1}$
\end{algorithmic}
\end{algorithm}

\begin{table}[!htbp]
\centering
\small
\renewcommand{\arraystretch}{1.1}
\begin{tabular*}{1\linewidth}{@{\extracolsep{\fill}}llll@{}}
\toprule
\parbox[t]{0.24\linewidth}{\raggedright\strut Route} &
\parbox[t]{0.36\linewidth}{\raggedright\strut Role of retrieved skill} &
\parbox[t]{0.10\linewidth}{\centering\strut Executes\\code} &
\parbox[t]{0.10\linewidth}{\centering\strut Indirect\\effect} \\
\midrule
\parbox[t]{0.24\linewidth}{\raggedright\strut Dynamic generation} &
\parbox[t]{0.36\linewidth}{\raggedright\strut No retrieved skill is used.} &
\parbox[t]{0.10\linewidth}{\centering\strut \xmark} &
\parbox[t]{0.10\linewidth}{\centering\strut \xmark} \\
\addlinespace[4pt]
\parbox[t]{0.24\linewidth}{\raggedright\strut Direct skill execution} &
\parbox[t]{0.36\linewidth}{\raggedright\strut A retrieved skill is called as an executable subroutine before verification.} &
\parbox[t]{0.10\linewidth}{\centering\strut \cmark} &
\parbox[t]{0.10\linewidth}{\centering\strut \xmark} \\
\addlinespace[4pt]
\parbox[t]{0.24\linewidth}{\raggedright\strut Skill-conditioned\\generation} &
\parbox[t]{0.36\linewidth}{\raggedright\strut Retrieved skills are supplied only as structured context for planning and generation.} &
\parbox[t]{0.10\linewidth}{\centering\strut \xmark} &
\parbox[t]{0.10\linewidth}{\centering\strut \cmark} \\
\bottomrule
\end{tabular*}
\caption{Routing modes in FlowEvo (Section~\ref{subsec:planning}).}
\label{tab:planning_modes}
\end{table}

\section{LLM Prompts}
\label{app:prompts}

This appendix lists the prompts used by the FlowEvo pipeline on ALFWorld and on the code/math benchmarks. Placeholders in \texttt{\{curly\_braces\}} denote runtime substitutions.

\subsection{ALFWorld action generation}
\label{app:prompt-alfworld-step}

Per-step action generation in \texttt{AlfWorldGenerator.step} (\texttt{src/alfworld\_/generator.py:69}), with \texttt{temperature=0.0}, \texttt{max\_output\_tokens=256}. The system prompt is fixed; the user prompt carries the observation, action history, and any retrieved skill context.

\begin{minipage}{\linewidth}
\begin{lstlisting}[language=PromptText,caption={ALFWorld step system prompt.}]
You are an expert household robot completing tasks in a virtual home. You will be given a task goal, the current observation, and a list of valid actions.

At each step:
1. Think about what you need to do next and why.
2. Choose exactly ONE action from the valid actions list.

Common task patterns:
- pick_and_place: go to object location -> take it -> go to destination -> put it
- pick_clean_then_place: go to object -> take -> go to sinkbasin -> clean -> go to dest -> put
- pick_heat_then_place: go to object -> take -> go to microwave -> heat -> go to dest -> put
- pick_cool_then_place: go to object -> take -> go to fridge -> cool -> go to dest -> put
- examine_in_light: go to object -> take -> go to lamp -> use lamp
- pick_two: find first object -> take -> go to dest -> put -> find second -> take -> go to dest -> put

Format your response as:
Think: <your step-by-step reasoning>
Act: <the exact action from the valid actions list>
\end{lstlisting}
\end{minipage}

\begin{minipage}{\linewidth}
\begin{lstlisting}[language=PromptText,caption={ALFWorld step user prompt template. The \texttt{\{skill\_context\}} block is empty in dynamic mode and carries retrieved skill content such as exemplars, insights, or the full template in skill-conditioned mode and its seeded variants.}]
Task: {task_goal}

Relevant experience:
{skill_context}

Recent actions:
  > {action_history[i]}
    {observation_history[i][:120]}
  ...

Current observation:
{observation}

Valid actions ({len}):
  {cmd_1}
  {cmd_2}
  ...

Choose ONE action from the valid actions list above:
\end{lstlisting}
\end{minipage}

\subsection{ALFWorld trace-to-skill compilation (Layer 2: guideline)}
\label{app:prompt-alfworld-compile}

Invoked by \texttt{AlfWorldCompiler.\_compile\_guideline} (\texttt{src/alfworld\_/compiler.py:200}) on successful traces. Settings: \texttt{temperature=0.0}, \texttt{max\_output\_tokens=500}.

\begin{minipage}{\linewidth}
\begin{lstlisting}[language=PromptText,caption={Guideline compilation system prompt.}]
You are a task analysis assistant. Extract concise procedural rules.
\end{lstlisting}
\end{minipage}

\begin{minipage}{\linewidth}
\begin{lstlisting}[language=PromptText,caption={Guideline compilation user prompt template. Abstractness constraints (no object names, no numbers, generic terms) are enforced in the prompt itself rather than by any separate post-processing step after generation.}]
You are analyzing a successful task execution trace to extract reusable procedural rules.

Task type: {task_type}
Task goal: {goal}

Successful action sequence:
  1. {action} -> {observation[:60]}
  2. {action} -> {observation[:60]}
  ...

Extract 3-5 reusable procedural rules that would help an agent solve similar tasks in the future.

Requirements:
- Each rule must be ABSTRACT: do NOT include specific object names, location names, or numbers
- Use generic terms like "target object", "cleaning station", "destination receptacle"
- Each rule should be one sentence in the format: "When [situation], [do what], because [why]"
- Focus on STRATEGY (search approach, action ordering) not exact steps
- Include at least one rule about what to AVOID (common failure patterns)
- 3-5 rules only, keep each concise

Rules:
\end{lstlisting}
\end{minipage}

\subsection{Code and math generation with skill context}
\label{app:prompt-codemath}

Primary generation in the code/math runner (\texttt{src/code\_math/runner.py:474}) is a single call parameterized by \texttt{\_GEN\_SETTINGS}\,$=$\,\texttt{(temperature=0.0, max\_output\_tokens=2048)} at Level~1 and \texttt{\_GEN\_L2}\,$=$\,\texttt{(0.5, 2048)} for the 2-candidate mini-ensemble at Level~2. Skill-conditioned generation uses the same call with a retrieved-skill block prepended to the user prompt, formatted as the bracketed reference prefix below:

\begin{minipage}{\linewidth}
\begin{lstlisting}[language=PromptText,caption={Code task skill-context prefix (prepended to the standard HumanEval problem statement). For math tasks the prefix reads ``Here is a similar solved problem for reference:''.}]
[Reference from similar solved problem:
{skill_context}
]
\end{lstlisting}
\end{minipage}

\section{Framework Thresholds and Configuration Constants}
\label{app:constants}

FlowEvo's routing, admission, and curation decisions are governed by a set of numeric thresholds. Table~\ref{tab:thresholds} lists the values that affect the results reported in Section~\ref{sec:exp}. The full set of 177 constants appears in the released code.

\begin{table}[!htbp]
\centering
\small
\renewcommand{\arraystretch}{1.1}
\begin{tabular*}{\linewidth}{@{\extracolsep{\fill}} l l @{}}
\toprule
Constant & Value \\
\midrule
\multicolumn{2}{@{}l}{\textit{Retrieval}} \\
\quad \texttt{top\_k}                                    & 3 \\
\quad \texttt{retrieval\_threshold}                      & 5.0 \\
\quad Historical positive-transfer cap                   & 2.5 \\
\quad Negative-transfer risk cap                         & 3.0 \\
\addlinespace[3pt]
\multicolumn{2}{@{}l}{\textit{Routing}} \\
\quad \texttt{\_COMPATIBILITY\_DIRECT\_THRESHOLD}        & 0.55 \\
\quad Direct-execute compat gate                         & $\ge 1.5$ \\
\quad \texttt{\_CODE\_EXCERPT\_THRESHOLD}                & 4.6 \\
\quad Max seed context skills                            & 2 \\
\addlinespace[3pt]
\multicolumn{2}{@{}l}{\textit{Admission}} \\
\quad \texttt{banned\_imports}                           & \texttt{\{os, subprocess, socket, requests, pathlib\}} \\
\quad \texttt{banned\_calls}                             & \texttt{\{eval, exec, compile, \_\_import\_\_, open\}} \\
\quad \texttt{prune\_audit\_fail\_threshold}             & 2 \\
\addlinespace[3pt]
\multicolumn{2}{@{}l}{\textit{Curation (ALFWorld)}} \\
\quad \texttt{\_CONTRASTIVE\_MIN\_GUIDED}                & 5 \\
\quad \texttt{\_CONTRASTIVE\_MIN\_UNGUIDED}              & 3 \\
\quad \texttt{\_CONTRASTIVE\_HARM\_THRESHOLD}            & $-0.1$ \\
\quad \texttt{\_AUDIT\_MIN\_USES}                        & 3 \\
\quad \texttt{\_AUDIT\_UTILITY\_THRESHOLD}               & 0.5 \\
\quad \texttt{\_AUDIT\_INACTIVITY\_LIMIT}                & 50 episodes \\
\addlinespace[3pt]
\multicolumn{2}{@{}l}{\textit{Adaptive escalation}} \\
\quad Level 1 (greedy)                                   & \texttt{temp=0.0, max=2048} \\
\quad Level 2 (mini-ensemble, 2 candidates)              & \texttt{temp=0.5, max=2048} \\
\quad Level 3 (retry, hot)                               & \texttt{temp=0.7, max=2048} \\
\quad Level 4 (retry, cool)                              & \texttt{temp=0.2, max=2048} \\
\addlinespace[3pt]
\multicolumn{2}{@{}l}{\textit{Execution}} \\
\quad ALFWorld \texttt{max\_steps}                       & 50 \\
\quad Sandbox timeout (code/math)                        & 10\,s \\
\quad Draft temperature / max tokens                     & 0.2 / 900 \\
\quad Repair temperature / max tokens                    & 0.15 / 900 \\
\bottomrule
\end{tabular*}
\caption{Core decision thresholds used in FlowEvo. All values are defaults; no per-benchmark tuning was performed for the reported results.}
\label{tab:thresholds}
\end{table}

The contrastive harm threshold $-0.1$, combined with the minimum-samples gates (\texttt{\_CONTRASTIVE\_MIN\_GUIDED}\,$=5$, \texttt{\_CONTRASTIVE\_MIN\_UNGUIDED}\,$=3$), determines when skill curation suppresses a skill: a candidate is suppressed only after at least 5 guided and 3 unguided samples, and only if the delta falls below $-0.1$. The suppression event reported in Section~\ref{subsec:ablation} (guided $2/14$, unguided $1/3$, $\Delta = -0.19$) satisfies these gates: $14 \ge 5$, $3 \ge 3$, and $0.143 - 0.333 = -0.19 < -0.1$. The analogous event in the representative run of Appendix~\ref{app:routing} (guided $1/10$ vs.\ unguided $1/3$, $\Delta = -0.23$) satisfies them as well: $10\ge 5$, $3\ge 3$, and $0.1-0.333 = -0.233 < -0.1$, so the suppression trigger fires in both runs.

\section{Skill Record Example and Bank Composition}
\label{app:skill-record}

This appendix shows a complete active skill record from a representative ALFWorld run (the run plotted in Figures~\ref{fig:dynamics} and~\ref{fig:per_task}; its overall success rate of 82.8\% lies within the 3-seed band $85.6 \pm 3.8$ reported in the main text), followed by the final composition of the skill bank in that run. The implementation organizes each skill record into 3 layers: a \emph{Layer-1 template} that captures the parameterized action sequence (Listing~\ref{lst:skill-template}), a \emph{Layer-2 guideline} produced by the compilation prompt of Appendix~\ref{app:prompt-alfworld-compile}, and a \emph{Layer-3 insight} that aggregates environment priors across traces of the same task type (Listing~\ref{lst:skill-insight}). We show the Layer-1 and Layer-3 content here; Layer-2 guidelines are generated directly from the prompt in Appendix~\ref{app:prompt-alfworld-compile} and are not reproduced a second time within this appendix.

\subsection{A compiled skill: \texttt{pick\_heat\_then\_place\_in\_recep}}
\label{app:skill-pickheat}

Key FlowEvo-specific fields in the record below: \texttt{action\_template} is the parameterized action sequence compiled from the source trace, with slots bound at runtime from goal parsing; \texttt{utility} is \texttt{success\_count / use\_count} (no smoothing); \texttt{status} $\in \{\texttt{active}, \texttt{suppressed}\}$ is set by curation; \texttt{contrastive\_stats} tracks a matched subset of guided vs.\ unguided episodes used for harm-delta estimation, and its counts are therefore typically smaller than \texttt{use\_count}. The insight record that follows the template is retrieved jointly with it and injected as structured context when direct execution is not selected by the router.

\begin{minipage}{\linewidth}
\begin{lstlisting}[language=PromptText,label=lst:skill-template,caption={Layer-1 template record for \texttt{pick\_heat\_then\_place\_in\_recep}. This is the on-disk representation used at retrieval time by the router.}]
{
  "template": {
    "template_id": "alf_skill_pick_heat_then_place_in_recep_112be592",
    "task_type": "pick_heat_then_place_in_recep",
    "action_template": [
      "go to {object_location}",
      "take {object} from {object_location}",
      "go to {heating_station}",
      "heat {object} with {heating_station}",
      "go to {target_location}",
      "move {object} to {target_location}"
    ],
    "parameters": ["heating_station", "object",
                   "object_location", "target_location"],
    "example_bindings": {
      "heating_station": "microwave 1",
      "object": "cup 1",
      "object_location": "cabinet 3",
      "target_location": "coffeemachine 1"
    },
    "source_task_id": "eval_out_of_distribution_16",
    "source_trace_id": "alf_ff42ccc3",
    "success_count": 1,
    "failure_count": 0
  },
  "utility": 1.0,
  "status": "active",
  "use_count": 21,
  "success_count": 21,
  "failure_count": 0,
  "last_used_episode": 134,
  "contrastive_stats": {
    "guided_success": 2, "guided_total": 2,
    "unguided_success": 3, "unguided_total": 3
  },
  "injection_suppressed": false
}
\end{lstlisting}
\end{minipage}

\begin{minipage}{\linewidth}
\begin{lstlisting}[language=PromptText,label=lst:skill-insight,caption={Layer-3 insight record accompanying the template above. Insights are generated after \texttt{\_INSIGHT\_MIN\_SAMPLES}\,$=3$ traces of the same task type and provide environment priors that generalize across parametric variation.}]
{
  "task_type": "pick_heat_then_place_in_recep",
  "search_priority": "countertop > cabinet > coffeemachine > garbagecan > microwave",
  "common_objects": [],
  "common_locations": ["countertop", "cabinet",
                       "coffeemachine", "garbagecan", "microwave"],
  "common_pitfalls": ["100% of failures are timeouts -- explore more efficiently"],
  "environment_facts": ["microwave is usually numbered 1"],
  "sample_count": 23
}
\end{lstlisting}
\end{minipage}

The insight is aggregated over \emph{all} \texttt{sample\_count}\,$=23$ traces of this task type, including the one \texttt{pure\_dynamic} episode that failed (see Table~\ref{tab:routing}). The \texttt{common\_pitfalls} field therefore describes a failure observed in the non-skill path; it is consistent with the Layer-1 template record reporting zero failures over its own 21 uses.

\subsection{Final skill-bank composition}
\label{app:skill-bank}

Table~\ref{tab:skill-bank} summarizes the bank at the end of the 134-episode representative run. The run compiles 6 templates, one per task type; the \texttt{pick\_two\_obj\_and\_place} template is suppressed by contrastive curation as discussed in Section~\ref{subsec:ablation}.

\begin{table}[!htbp]
\centering
\small
\renewcommand{\arraystretch}{1.1}
\begin{tabular*}{\linewidth}{@{\extracolsep{\fill}} l c c c c c @{}}
\toprule
Task type & Status & Uses & Success & Fail & Utility \\
\midrule
\texttt{pick\_and\_place\_simple}              & active      & 23 & 23 & 0 & 1.00 \\
\texttt{pick\_clean\_then\_place\_in\_recep}   & active      & 28 & 28 & 0 & 1.00 \\
\texttt{pick\_heat\_then\_place\_in\_recep}    & active      & 21 & 21 & 0 & 1.00 \\
\texttt{pick\_cool\_then\_place\_in\_recep}    & active      & 20 & 18 & 2 & 0.90 \\
\texttt{look\_at\_obj\_in\_light}              & active      & 15 & 13 & 2 & 0.87 \\
\texttt{pick\_two\_obj\_and\_place}            & suppressed  & 10 &  1 & 9 & 0.10 \\
\bottomrule
\end{tabular*}
\caption{Skill bank after 134 ALFWorld episodes (representative run). ``Uses'' counts episodes in which the skill was retrieved above \texttt{retrieval\_threshold} and thus participated in routing (whether the router eventually selected direct execution or skill-conditioned generation, or downgraded to dynamic). ``Success''\,/\,``Fail'' are the outcomes of those retrieval-matched episodes; ``Utility'' $=$ \texttt{Success / Uses}. Entries are cumulative over the run. For 5 of the 6 task types, Uses equals \texttt{Direct}$+$\texttt{Cond} from Table~\ref{tab:routing}; \texttt{look\_at\_obj\_in\_light} has 3 additional retrieval matches that were routing-downgraded to \texttt{pure\_dynamic} (see Section~\ref{subsec:analysis}: ``the skill matched on 15 of 18 episodes'').}
\label{tab:skill-bank}
\end{table}

The \texttt{pick\_two\_obj\_and\_place} row reflects the state at the end of the run: 10 guided episodes accumulated before suppression (Cond\,$=10$ in Table~\ref{tab:routing}), with one success and 9 failures, yielding a final utility of $1/10 = 0.10$. After suppression, subsequent \texttt{pick\_two\_obj\_and\_place} episodes bypassed the bank entirely and are \emph{not} counted in this row (they contribute to the Dyn column of Table~\ref{tab:routing}).

\section{Per-Task-Type Routing Distribution}
\label{app:routing}

Table~\ref{tab:routing} breaks down the 134 ALFWorld episodes of the representative run by the router decision actually taken. It distinguishes 4 internal modes: \texttt{direct\_skill} (template called as an executable subroutine, no fallback), \texttt{direct\_then\_*} (template invoked but execution required fallback to skill-conditioned or dynamic generation), \texttt{exemplar\_guided} (skill supplied as context only), and \texttt{pure\_dynamic} (no retrieved skill is used).

\noindent\textbf{Terminology mapping to the main text.}
The main text uses a coarser classification:
\begin{itemize}[leftmargin=*,noitemsep,topsep=2pt]
  \item ``Direct-reuse route'' $=\texttt{direct\_skill} \cup \texttt{direct\_then\_*}$: 101/134 episodes with 99 successes (98\%), as reported in Section~\ref{subsec:analysis} and plotted as the reuse curve in Figure~\ref{fig:dynamics}(b).
  \item ``Skill-conditioned generation'' $=\texttt{exemplar\_guided}$ and related skill-context-only variants: retrieved skills supplied as structured context without template invocation.
  \item ``Dynamic generation'' $=\texttt{pure\_dynamic}$: no retrieved skill is used.
\end{itemize}
In Table~\ref{tab:routing}, the Direct column reports \texttt{direct\_skill} only; the Cond.\ column collapses \texttt{direct\_then\_*} and \texttt{exemplar\_guided} together. The paper's ``direct-reuse'' count of 101 therefore equals Direct (67) $+$ the \texttt{direct\_then\_*} subset of Cond.\ (34), rather than the narrower Direct column alone, which would understate reuse.

\begin{table}[!htbp]
\centering
\small
\renewcommand{\arraystretch}{1.1}
\begin{tabular*}{\linewidth}{@{\extracolsep{\fill}} l c c c c c c @{}}
\toprule
Task type & $n$ & Direct & Cond. & Dyn. & Direct SR & Overall SR \\
\midrule
\texttt{look\_at\_obj\_in\_light}              & 18 &  0 & 12 &  6 & ---      & 14/18 (78\%) \\
\texttt{pick\_and\_place\_simple}              & 24 & 14 &  9 &  1 & 14/14    & 24/24 (100\%) \\
\texttt{pick\_clean\_then\_place\_in\_recep}   & 31 & 22 &  6 &  3 & 22/22    & 29/31 (94\%) \\
\texttt{pick\_cool\_then\_place\_in\_recep}    & 21 & 12 &  8 &  1 & 12/12    & 19/21 (90\%) \\
\texttt{pick\_heat\_then\_place\_in\_recep}    & 23 & 19 &  2 &  2 & 19/19    & 22/23 (96\%) \\
\texttt{pick\_two\_obj\_and\_place}            & 17 &  0 & 10 &  7 & ---      &  3/17 (18\%) \\
\midrule
Total                                 & 134 & 67 & 47 & 20 & 67/67    & 111/134 (83\%) \\
\bottomrule
\end{tabular*}
\caption{Internal routing modes by task type on ALFWorld (representative run). ``Direct'' $=$ pure \texttt{direct\_skill}; ``Cond.'' $=$ \texttt{direct\_then\_*} plus \texttt{exemplar\_guided}; ``Dyn.'' $=$ \texttt{pure\_dynamic}. ``Direct SR'' is the success rate among pure direct-execute episodes. ``Overall SR'' is the success rate on all episodes of the task type.}
\label{tab:routing}
\end{table}

\noindent\textbf{Curation suppression event.}
One suppression event was triggered during the run, satisfying the curation gates of Appendix~\ref{app:constants}:
\begin{itemize}[leftmargin=*,noitemsep,topsep=2pt]
  \item \texttt{pick\_two\_obj\_and\_place}: guided 1/10 (10\%) vs.\ unguided 1/3 (33\%), $\Delta = -0.23$; template suppressed (the event in one seed of the 3-seed campaign, $\Delta = -0.19$ over 14 guided episodes, is analyzed in Section~\ref{subsec:ablation} with the suppression timeline).
\end{itemize}
After suppression, the remaining 4 \texttt{pick\_two\_obj\_and\_place} episodes were routed through \texttt{pure\_dynamic} (contributing to the Dyn count of $7 = 3$ pre-suppression $+\,4$ post-suppression in Table~\ref{tab:routing}); they did not use the skill bank and are therefore excluded from the \texttt{Uses} count reported in Table~\ref{tab:skill-bank} for the corresponding row.

\section{LLM Usage Disclosure}
\label{app:llm-usage}

LLMs are the system under study: every FlowEvo component and every baseline runs on the backbone models specified in Section~\ref{subsec:setup}. No LLM was used as a judge for any reported metric; all accuracy numbers are computed programmatically. During writing, Claude Fable~5 (Anthropic) was used to polish the text (grammar, wording, and sentence-level edits). The research ideas, method, experiments, and analyses are the authors' own. No experimental data or results were generated by an LLM, and the authors reviewed every edit and take full responsibility for the entire content of the paper.

\end{document}